\documentclass{article}

\usepackage{microtype}
\usepackage{graphicx}
\usepackage{subcaption}
\usepackage{enumitem}
\usepackage{booktabs} 
\usepackage{siunitx}
\usepackage{multirow}
\usepackage{makecell}
\usepackage{xcolor}
\usepackage{colortbl}
\usepackage{listings}
\usepackage[table]{xcolor}
\usepackage[most]{tcolorbox}

\definecolor{hl}{gray}{0.94}
\definecolor{modelbg}{RGB}{255,248,240}   
\definecolor{modelframe}{RGB}{153,76,0}   

\lstdefinestyle{pycode}{
    language=Python,
    basicstyle=\ttfamily\small,
    keywordstyle=\color{blue},
    commentstyle=\color{green!50!black},
    stringstyle=\color{red!70!black},
    showstringspaces=false,
    breaklines=true,
    tabsize=4
}

\usepackage{hyperref}

\usepackage[accepted]{icml2026}

\usepackage{amsmath}
\usepackage{amssymb}
\usepackage{mathtools}
\usepackage{amsthm}

\usepackage[capitalize,noabbrev]{cleveref}

\theoremstyle{plain}

\theoremstyle{definition}

\theoremstyle{remark}

\usepackage[textsize=tiny]{todonotes}

\begin{document}

\twocolumn[
  \icmltitle{Strategy-Aware Optimization Modeling with Reasoning LLMs}

  \icmlsetsymbol{equal}{*}
  \icmlsetsymbol{corr}{\textdagger}
  
  \begin{icmlauthorlist}
    \icmlauthor{Ruiqing Zhao}{sch1,equal}
    \icmlauthor{Fengzhi Li}{comp,equal}
    \icmlauthor{Yuan Zuo}{sch2,corr}
    \icmlauthor{Rui Liu}{sch1,corr}
    \icmlauthor{Yansong Liu}{sch1}
    \icmlauthor{Yunfei Ma}{comp}
    \icmlauthor{Fanyu Meng}{comp}
    \icmlauthor{Junlan Feng}{comp}
  \end{icmlauthorlist}

  \icmlaffiliation{sch1}{School of Computer Science and Engineering, Beihang University, Beijing, China}
  \icmlaffiliation{sch2}{MIIT Key Laboratory of Data and Decision Intelligence, Beihang University, Beijing, China}
  \icmlaffiliation{comp}{JIUTIAN Research, Beijing, China}

  \icmlcorrespondingauthor{Yuan Zuo}{zuoyuan@buaa.edu.cn}
  \icmlcorrespondingauthor{Rui Liu}{lr@buaa.edu.cn}

  \vskip 0.3in
]

\printAffiliationsAndNotice{\icmlEqualContribution}

\begin{abstract}
Large language models (LLMs) can generate syntactically valid optimization programs, yet often struggle to reliably choose an effective modeling strategy, leading to incorrect formulations and inefficient solver behavior. We propose \textsc{SAGE}, a strategy-aware framework that makes \emph{Modeling Strategy} explicit in both data construction and post-training. \textsc{SAGE} builds a solver-verified multi-strategy dataset and trains a student model with supervised fine-tuning followed by Segment-Weighted GRPO using a composite reward over format compliance, correctness, and solver efficiency. Across eight benchmarks spanning synthetic and real-world settings, \textsc{SAGE} improves average pass@1 from 72.7 to 80.3 over the strongest open-source baseline. With multiple generations, \textsc{SAGE} discovers more distinct correct formulations and improves component-level diversity at pass@16 by 19--29\%. At the largest scale, \textsc{SAGE} produces more compact constraint systems with 14.2\% fewer constraints than the baseline, consistent with solver-efficient modeling. Overall, these results show that making \emph{Modeling Strategy} explicit improves automated optimization modeling. Code is available at https://github.com/rachhhhing/SAGE.
\end{abstract}

\section{Introduction}

In operations research (OR), optimization models translate informal decision-making requirements into precise mathematical programs that can be solved by modern optimization software.
In many practical workflows, however, the bottleneck is not running a solver, but formulating a solver-executable model from a natural-language description.
This step requires careful design of index sets, decision variables, and constraints, and ideally produces formulations that are not only correct but also efficient to solve \cite{antoniou2007practical}.
The gap between problem intent and solver-ready formulations remains a major source of cost across scheduling, manufacturing, logistics, and finance \cite{cohen2005scheduling,jayal2010manufacturing,bartolacci2012logistics,ponsich2012finance}.

Large language models (LLMs) have recently demonstrated strong capabilities in language understanding, reasoning, and code generation \cite{brown2020language}, spurring rapid progress on automated optimization modeling.
Existing systems generally follow two lines:
prompt- and agent-based approaches decompose a problem and synthesize solver code \cite{xiao2023chainofexperts,ahmaditeshnizi2024optimus,zhang2025orllmagent}, while learning-based methods fine-tune models on optimization data to improve formulation accuracy and executability \cite{huang2025orlm,jiang2024llmopt,chen2025sirl,zhou2025steporlm}.
Despite steady improvements, a recurring limitation is that key formulation decisions remain implicit: models are typically trained to generate mathematical formulations and solver code, but without explicit globally design that guide how a problem should be modeled.

In practice, optimization modeling is \emph{strategy-driven}.
Before writing down variables and constraints, human experts typically commit to a high-level formulation paradigm such as flow-based or assignment-based modeling, which determines the decision domain and the variable backbone that structures the entire model \cite{williams2013model,hillier2005introduction}.
We call this paradigm-level choice the \textit{Modeling Strategy}.
It is not merely stylistic.
It enforces global consistency among variables, constraints, and objectives, and it strongly shapes solver behavior in practice, since solution effort can depend critically on the chosen formulation \cite{lindo_formulation_time}.

\begin{figure*}[t]
  \centering
  \includegraphics[width=0.85\linewidth]{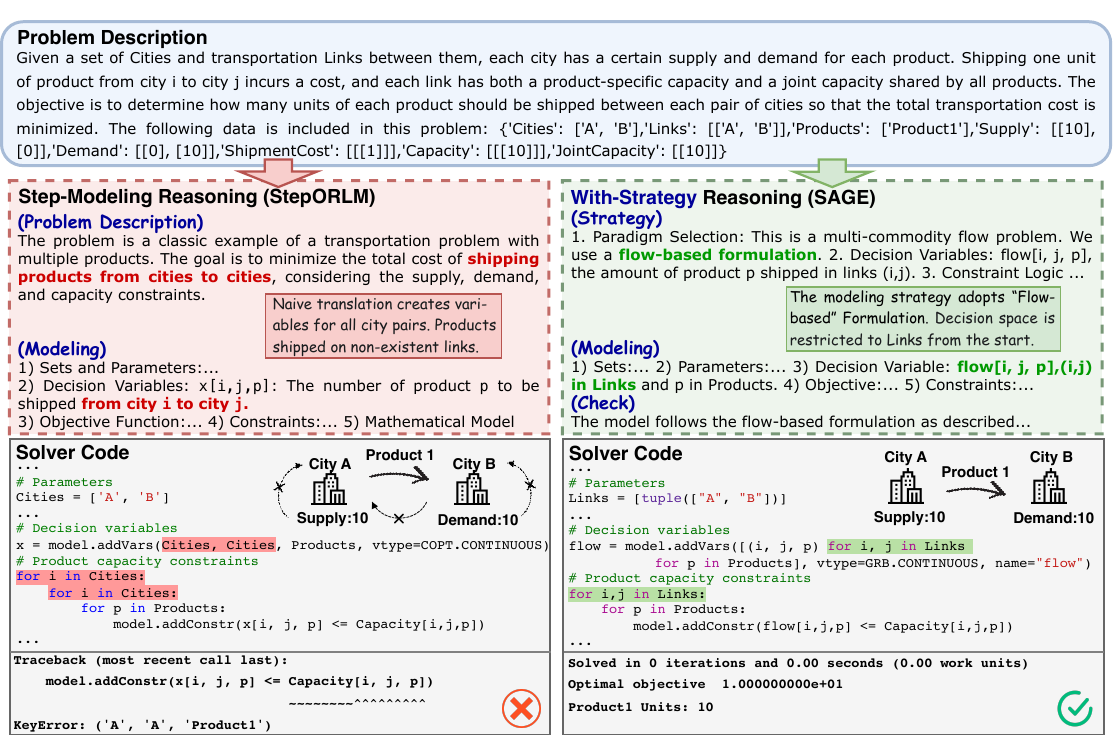}
  \caption{\textbf{Why modeling strategy matters.} A step-wise pipeline may define variables on an incorrect index space (e.g., $(A,A)$), creating invalid arcs and runtime failures (e.g., \texttt{KeyError}). Strategy-aware reasoning first commits to a paradigm (e.g., flow-based) and restricts the decision domain (e.g., \texttt{Links}), producing a consistent and solver-executable model.}
  \label{fig:case}
  \vspace{-0.2cm}
\end{figure*}

Figure~\ref{fig:case} shows a common failure mode through a simple transportation scenario. In this problem, goods can only be shipped along a predefined set of transportation links between cities. However, a step-wise modeling pipeline may blindly introduce shipment variables for all city pairs, implicitly assuming that goods can be sent between any two cities. This creates decision variables for non-existent routes (e.g., $(A,A)$) and and it leads to runtime failures.
In contrast, a strategy-aware approach commits early to a flow-based formulation and restricts variables to $(i,j)\in$\texttt{Links}, yielding a globally consistent, solver-executable model.
This highlights the first role of modeling strategy: ensuring \emph{executability and correctness} via coherent decision domains.

Strategy also matters for \emph{efficiency}.
Many OR problems admit multiple correct formulations under different paradigms, yet they can differ substantially in model size, constraint tightness, and solver performance \cite{calafiore2014optimization}.
Thus, an automation system should learn not only to generate correct and executable formulations, but also to choose strategies that lead to efficient solves (e.g., reduced solve time).
This remains challenging because (i) strategy-level reasoning is rarely explicit in training data, making paradigm selection hard to learn; and (ii) prevailing objectives emphasize executability and correctness but provide limited signal for efficiency-aware strategy learning.

\begin{figure*}[t]
  \centering
  \includegraphics[width=0.9\linewidth]{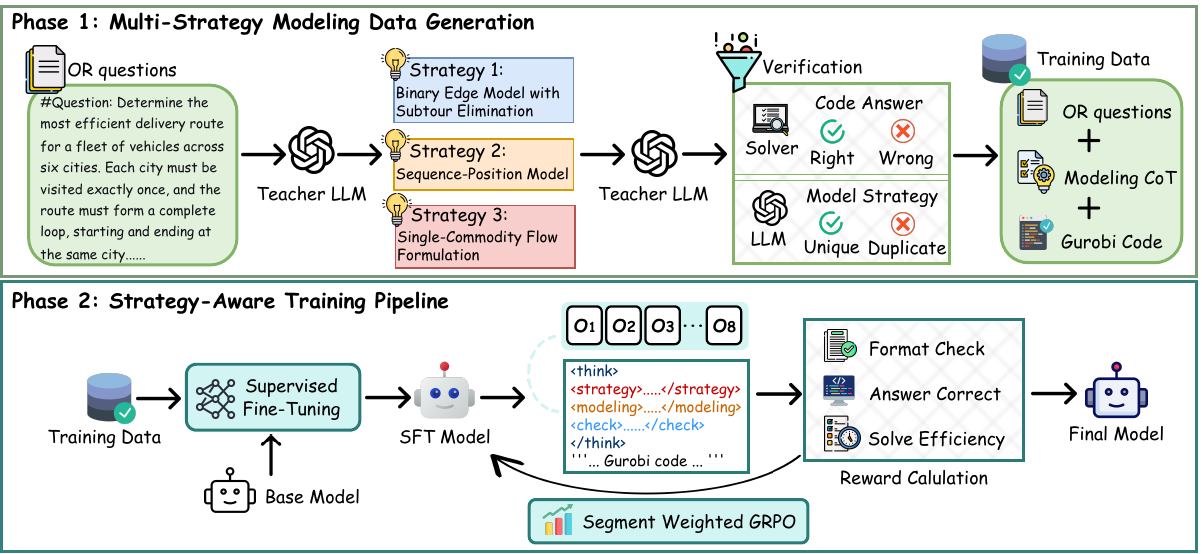}
  \caption{\textbf{Overview of \textsc{SAGE}.} Phase~1 builds a multi-strategy, solver-verified corpus by generating multiple candidate strategies per problem, producing strategy-conditioned reasoning and Gurobi code, filtering via solver validation against ground-truth, and deduplicating redundant strategies with an LLM-as-Judge. Phase~2 trains with supervised fine-tuning and Segment-Weighted GRPO using format, correctness, and efficiency rewards.}
  \label{fig:overview}
\end{figure*}

To address these challenges, we study \emph{Strategy-Aware Optimization Modeling with Reasoning LLMs} and present \textsc{SAGE} (\textbf{S}trategy-\textbf{A}ware \textbf{G}uided r\textbf{E}asoning), a generation-based framework for optimization modeling with reasoning LLMs (Figure~\ref{fig:overview}).
\textsc{SAGE} makes modeling strategy explicit in both data construction and post-training.
In Phase~1, a teacher model generates multiple candidate modeling strategies for each OR problem and then produces a reasoning trajectory together with the corresponding Gurobi code conditioned on each strategy,.
We execute the code and validate solutions against the ground-truth answer to filter incorrect outputs, and deduplicate semantically redundant strategies via an LLM-as-Judge to retain distinct ones.
In Phase~2, we optimize the model with solver feedback via \emph{Segment-Weighted GRPO}, which improves credit assignment by emphasizing strategy decisions, and uses a composite reward that jointly captures (i) structured format compliance, (ii) executability and correctness, and (iii) solver efficiency.
Our contributions are:
\begin{itemize}[topsep=0pt, itemsep=2pt, parsep=0pt, partopsep=0pt]
    \item We construct a solver-verified, multi-strategy dataset of (question, reasoning, code) by generating multiple strategies per problem, filtering outputs via solver validation against ground-truth, and deduplicating redundant strategies.
    \item We propose \emph{Segment-Weighted GRPO} that upweights strategy segments and optimizes a composite reward over format compliance, executability and correctness, and solver efficiency.
    \item Across eight benchmarks, \textsc{SAGE} improves average pass@1 from 72.7 to 80.3 over the strongest open-source baseline, while discovering more distinct correct formulations under pass@K and producing more solver-efficient models.
\end{itemize}

\section{Methodology}

We propose SAGE, a two-phase framework that makes \emph{modeling strategy} explicit throughout. In Phase1, we perform solver-verified multi-strategy data synthesis to produce \textit{(question, modeling reasoning, code)} triples. In Phase2, we post-train the model using supervised fine-tuning (SFT) followed by reinforcement learning (RL) with solver feedback to optimize structured reasoning, executability and correctness, and solver efficiency.

\subsection{Multi-Strategy Modeling Data Synthesis}
Existing optimization-modeling datasets often provide only \textit{(question, model, code)} sequences, which expose final formulations but omit the strategy-level reasoning process. This makes it difficult for models to learn how to \emph{choose} a formulation paradigm and consistently instantiate it. We therefore construct a dataset of \textit{(question, modeling reasoning, code)} triples, where the \textit{modeling reasoning} explicitly separates (i) paradigm selection from (ii) formulation instantiation.

Given a set of optimization problems $\mathcal{Q}=\{q_i\}_{i=1}^{N}$ with ground-truth solver answers, a strategy teacher first proposes $K$ candidate modeling strategies for each problem:
\begin{equation}
\mathcal{S}(q_i)=\{s_i^{(k)}\}_{k=1}^{K},
\end{equation}
where each $s_i^{(k)}$ corresponds to a distinct formulation paradigm (e.g., assignment-based vs.\ flow-based vs.\ time-indexed).

Then, a reasoning teacher model takes $(q_i, s_i^{(k)})$ as input and produces a complete reasoning trace and a solver-executable program:
\begin{equation}
(r_i^{(k)}, c_i^{(k)}) \sim \pi_{\text{teach}}(r, c \mid q_i, s_i^{(k)}),
\end{equation}
where $r_i^{(k)}$ describes the strategy-aligned modeling steps (sets/parameters/variables/constraints/objective) and $c_i^{(k)}$ is the corresponding Gurobi Python code.

\paragraph{Solver verification and filtering.}
To ensure data quality, we execute each generated program and validate it with the solver. We discard outputs that (i) fail to run, (ii) are infeasible or unbounded when a feasible optimum is expected, or (iii) return an solution that does not match the ground-truth answer within tolerance.

\paragraph{Strategy de-duplication.}
Multiple strategies may yield semantically equivalent formulations. We therefore apply an LLM-as-Judge procedure~\cite{zheng2023judging} that jointly compares reasoning traces and code to remove semantically redundant strategies, retaining only distinct and meaningful paradigms.

The resulting supervised corpus is
\begin{equation}
\mathcal{D}_{\text{SFT}}=\bigcup_{q_i\in\mathcal{Q}}\{(q_i, r_i^{(k)}, c_i^{(k)}) \mid k\in\mathcal{K}_i\},
\end{equation}
where $\mathcal{K}_i$ indexes the remaining verified and de-duplicated strategies for problem $q_i$ (possibly multiple per problem). Concrete prompt templates and examples are provided in Appendix~\ref{app:prompts}.

\subsection{Supervised Fine-Tuning}
After constructing $\mathcal{D}_{\text{SFT}}$, we perform full-parameter supervised fine-tuning to \emph{distill} the teacher’s strategy-conditioned modeling reasoning and code generation behavior into the student model. Given a problem description $q$, the student model is trained to generate a reasoning trace $r$ followed by solver code $c$.

We concatenate the reasoning trace and solver code into a single target sequence $y=[r,c]$ and optimize the standard language-modeling objective:
\begin{equation}
\mathcal{L}_{\mathrm{SFT}}
= - \sum_{i=1}^{|\mathcal{D}_{\text{SFT}}|} \sum_{t=1}^{|y_i|}
\log P_\theta \big( y_{i,t} \mid q_i, y_{i,1:t-1} \big),
\end{equation}
where $y_{i,t}$ is the $t$-th token of the target sequence for sample $i$, and $\theta$ denotes model parameters.

\subsection{Reinforcement Learning with Solver Feedback}
\paragraph{Structured reasoning template.}
Unstructured ``one-block'' reasoning is often brittle for complex multi-stage tasks~\cite{wei2022chain,yao2023tree}. In optimization modeling, early paradigm decisions constrain all modeling steps, making explicit structure particularly important. We therefore enforce a template in which the model emits a \texttt{<think>} block with three ordered segments:
\begin{itemize}[topsep=0pt, itemsep=2pt, parsep=0pt, partopsep=0pt]
    \item \texttt{<strategy>}: identify the problem class, choose a modeling paradigm, and commit to the core decision variables and constraint logic;
    \item \texttt{<modeling>}: instantiate the formulation (sets, parameters, variables, objective, constraints) consistent with the chosen strategy;
    \item \texttt{<check>}: verify global consistency (index domains, loop dimensions, logical coupling) and remove obvious redundancies.
\end{itemize}
The solver-executable code is generated after the \texttt{<think>} block. This separation strengthens the linkage between high-level paradigm selection and concrete implementation, reducing index-space mismatches and other common errors.

\paragraph{Segment-weighted GRPO.}
For RL, we adopt Group Relative Policy Optimization (GRPO) \cite{shao2024deepseekmath}, a critic-free variant of Proximal Policy Optimization (PPO) \cite{schulman2017ppo}, which optimizes a PPO-style clipped objective using group-relative advantage estimates over multiple sampled completions for the same prompt. However, standard GRPO applies uniform token-level credit across the entire trajectory, implicitly assuming that all generation steps contribute equally to the final reward. This assumption is known to exacerbate the credit assignment problem in long-horizon reasoning tasks, where early high-level decisions may dominate downstream correctness, while later surface-level tokens have marginal impact \cite{lightman2023let, guo2025segment}. In optimization modeling, different reasoning stages play unequal roles in determining the correctness and efficiency of the final formulation.
We therefore introduce \emph{segment-weighted} GRPO, which assigns token weights according to the reasoning segment that generated them.

For a prompt, we sample a group of $G$ trajectories. Let $T_i$ be the trajectory length and $A_i$ the group-relative advantage. The segment-weighted GRPO objective is
\begin{equation}
\resizebox{0.98\linewidth}{!}{$
\mathcal{L}_{\mathrm{SW\_GRPO}}
= - \sum_{i=1}^{G} \sum_{t=1}^{T_i}
\alpha_t
\min\!\left(
\rho_{i,t} A_i,\;
\mathrm{clip}\!\left(\rho_{i,t},\, 1-\eta,\, 1+\eta\right) A_i
\right),
$}
\end{equation}
where $\rho_{i,t} = \frac{\pi_\theta(a_{i,t} \mid s_{i,t})}{\pi_{\theta_{\mathrm{old}}}(a_{i,t} \mid s_{i,t})}$ is the likelihood ratio and $\eta$ is the clipping parameter. The token weight $\alpha_t$ is chosen by segment membership, with
\begin{equation}
\alpha_{\mathrm{strategy}} > \alpha_{\mathrm{modeling}} > \alpha_{\mathrm{check}} > 0.
\end{equation}
Setting all weights to $1$ recovers standard GRPO.

\paragraph{Composite reward: format, outcome, and efficiency.}
We optimize not only for correctness but also for efficient formulations. For a generated response $y$, the total reward is
\begin{equation}
R(y)=R_{\mathrm{format}}(y)+R_{\mathrm{outcome}}(y)+R_{\mathrm{efficiency}}(y).
\end{equation}

\textbf{Format reward.}
We reward adherence to the structured template by checking the presence of the tags $\{\texttt{<think>},\texttt{<strategy>},\texttt{<modeling>},\texttt{<check>}\}$ and a final Python code block. Each correctly detected component contributes $0.2$, yielding
\begin{equation}
R_{\mathrm{format}}(y)=0.2 \times N_{\mathrm{seg}}(y),
\end{equation}
where $N_{\mathrm{seg}}(y)\in\{0,\dots,5\}$ is the number of correctly formatted components.

\textbf{Outcome reward.}
We execute the generated Gurobi program and assign
\begin{equation}
R_{\mathrm{outcome}}(y)=
\begin{cases}
0, & \text{if code execution fails},\\
0.2, & \text{if the solver reports infeasible},\\
0.4, & \text{if a solution returned is incorrect},\\
1.0, & \text{if the solution matches ground truth}.
\end{cases}
\end{equation}

\textbf{Efficiency reward.}
The efficiency reward is applied only when $R_{\mathrm{outcome}}(y)=1.0$. Let $M(y)$ be a normalized efficiency metric computed from solver feedback (e.g., solve time, iteration counts, or LP Gap); then
\begin{equation}
R_{\mathrm{efficiency}}(y)=1-\tanh\!\left(\frac{M(y)}{\alpha_{\mathrm{eff}}}\right),
\end{equation}
where $\alpha_{\mathrm{eff}}$ is a scaling constant. The $\tanh(\cdot)$ shaping keeps the reward in $[0,1]$ while providing smooth gradients. The detailed definition of $M(y)$ is provided in Appendix~\ref{app:eff_reward}.

\begin{table*}[t]
\centering
\caption{\textbf{The overall performance with Pass@1 accuracy (\%).}
All reproduced baseline results are obtained on the same cleaned benchmark versions used to evaluate SAGE.
Scores reported from original or reproduced papers are marked with ($^{*}$), 
while missing entries are denoted by (--). 
The best results are highlighted in \textbf{bold} and the second-best results are \underline{underlined}.}
\label{tab:opt_results}
\small
\setlength{\tabcolsep}{4pt}
\renewcommand{\arraystretch}{1.1}

\begin{tabular}{ll *{8}{S[table-format=2.1, detect-weight]} c}
\toprule
\multirow{2.5}{*}{Types} & \multirow{2.5}{*}{Models} & \multicolumn{4}{c}{Easy Tasks} & \multicolumn{4}{c}{Complex Tasks} & \multirow{2.5}{*}{Avg.} \\
\cmidrule(lr){3-6} \cmidrule(lr){7-10}
& & {NL4OPT} & {\makecell{MAMO\\Easy}} & {NLP4LP} & {OptiB.} & {\makecell{MAMO\\Cpx.}} & {CpxOR} & {IndOR} & {OptM.} & \\
\midrule

\multirow{5}{*}{Zero-shot}
& GPT-4o            & 89.2 & 77.2 & 89.9 & 82.9 & 61.3 & 50.0 & 47.6 & 21.1 & 64.9 \\
& DeepSeek-V3       & 87.8 & 95.2 & 87.6 & 85.1 & 55.9 & 55.6 & \underline{66.7} & \underline{40.4} & 71.8 \\
& DeepSeek-R1       & 86.4 & 88.1 & 81.5 & 77.4 & 63.9 & 55.6 & 57.1 & 34.9 & 68.1 \\
& Qwen3-32B         & 82.7 & 73.6 & 88.8 & 81.6 & 46.8 & 33.3 & 42.9 & 14.5 & 58.0 \\
& Qwen2.5-72B       & 84.0 & 91.9 & 85.4 & 80.4 & 52.3 & 44.4 & 47.6 & 18.1 & 63.0 \\
\midrule

\multirow{2}{*}{\makecell[l]{Agent-\\based}}
& Chain-of-Experts  & 66.7{$^*$} & 94.4{$^*$} & 87.4{$^*$} & 71.2{$^*$} & 50.6{$^*$} & \underline{57.1$^*$} & 31.2{$^*$} & {--} & {--} \\
& OptiMUS           & 76.2{$^*$} & 78.0{$^*$} & 88.8{$^*$} & 87.6{$^*$} & 46.8{$^*$} & 46.8{$^*$} & 45.2{$^*$} & 20.2{$^*$} & 61.2{$^*$} \\
\midrule

\multirow{3}{*}{\makecell[l]{Offline-\\learning}}
& ORLM-L3-8B        & 73.8{$^*$} & 90.4{$^*$} & 76.4{$^*$} & 61.8{$^*$} & 59.5{$^*$} & 50.0{$^*$} & 42.9{$^*$} & 2.6{$^*$} & 57.2{$^*$} \\
& LLMOpt-Q2.5-14B   & 80.3{$^*$} & 89.5{$^*$} & 73.4{$^*$} & 53.8{$^*$} & 44.1{$^*$} & 35.3{$^*$} & 29.0{$^*$} & 12.5{$^*$} & 52.2{$^*$} \\
& OptM-Q2.5-7B      & 94.7{$^*$} & 86.5{$^*$} & {--}        & 57.9{$^*$} & 51.2{$^*$} & {--}        & 20.0{$^*$} & 24.4{$^*$} & {--} \\
\midrule

\multirow{3}{*}{\makecell[l]{Online-\\RL}}
& SIRL-Q2.5-7B      & \underline{94.8} & \textbf{98.0} & 96.6 & \underline{89.6} & \underline{81.1} & 44.4 & 50.0 & 27.1 & \underline{72.7} \\
& StepORLM-Q3-8B    & \textbf{96.7} & \underline{95.4} & \underline{97.2} & 88.6 & 79.3 & 50.0 & 52.4 & 13.9 & 71.7 \\
\cmidrule(lr){2-11}
\rowcolor{hl}
\cellcolor{white} & \textbf{SAGE-DS-14B (Ours)} & 94.3 & 94.7 & \textbf{98.9} & \textbf{93.8} & \textbf{84.7} & \textbf{61.1} & \textbf{69.0} & \textbf{45.8} & \textbf{80.3} \\
\bottomrule
\end{tabular}
\vspace{0.25cm}
\end{table*}

\section{Experiments}
\subsection{Experiment Setup}

\paragraph{Benchmarks.}
We evaluate \textsc{SAGE} on a diverse set of optimization modeling benchmarks, including NL4OPT~\cite{ramamonjison2023nl4opt}, MAMO~\cite{huang2024mamo}, which is split into EasyLP and ComplexLP subsets, NLP4LP~\cite{ahmaditeshnizi2024optimus}, ComplexOR~\cite{xiao2023chainofexperts}, IndustryOR~\cite{huang2025orlm}, OptiBench~\cite{yang2024optibench}, and OptMATH~\cite{lu2025optmath}. Additional details for each benchmark are provided in Appendix~\ref{app:benchmark}.

We group these datasets into Easy and Complex categories. The Easy datasets include NL4OPT, MAMO Easy, NLP4LP, and OptiBench. They primarily contain classical linear programming problems with relatively few variables and constraints. The Complex datasets include MAMO Complex, ComplexOR, IndustryOR, and OptMATH. They cover a wider range of problem types and feature implicit or hierarchical constraints, as well as domain specific real world scenarios.

\paragraph{Baselines.}
We benchmark against a broad and representative set of baselines in four categories.
\begin{itemize}
    \item \textbf{Zero shot generalist LLMs}. Foundation models without OR specific training, including GPT-4o~\cite{hurst2024gpt}, DeepSeek-V3~\cite{liu2024deepseek}, DeepSeek-R1~\cite{guo2025deepseek}, Qwen3-32B~\cite{yang2024qwen2}, and Qwen2.5-72B-Instruct~\cite{yang2025qwen3}. All models are evaluated using the same prompts and decoding settings as our method.
    \item \textbf{Agent based methods}. Multi step orchestration frameworks that coordinate reasoning and code generation through explicit workflows. We include Chain-of-Experts~\cite{xiao2023chainofexperts} and OptiMUS~\cite{ahmaditeshnizi2024optimus}.
    \item \textbf{Offline learning baselines}. Supervised fine tuned models trained on optimization specific datasets, including ORLM-LLaMA-3-8B~\cite{huang2025orlm}, LLMOpt-Qwen2.5-14B~\cite{jiang2024llmopt}, and OptMATH-Qwen2.5-7B~\cite{lu2025optmath}.
    \item \textbf{Online RL baselines}. Models trained with reinforcement learning using solver based or process level feedback, including SIRL-Qwen2.5-7B~\cite{chen2025sirl} and StepORLM-Qwen3-8B~\cite{zhou2025steporlm}.
\end{itemize}

\paragraph{Implementation Details.}
We use DeepSeek-R1 as the teacher model during data construction and initialize our student model from DeepSeek-R1-distill-Qwen-14B.
Training problems are drawn from OptMATH-201K and IndustryOR-3K, which correspond to the training splits released by the original benchmarks.
This yields 100K samples for supervised fine tuning and an additional 8K samples for reinforcement learning.
We also repeat the same pipeline with a smaller Qwen3-8B backbone (initializing the student from Qwen3-8B) to verify the robustness across model scales; the corresponding results are reported in Appendix~\ref{app:res_qwen3}.
We use the VeRL framework~\cite{sheng2025hybridflow} with GRPO.
The detailed hyperparameter settings for both the supervised fine-tuning (SFT) and reinforcement learning (RL) stages are provided in Appendix~\ref{app:hyperparameter}.

\subsection{Main Results}
Table~\ref{tab:opt_results} reports Pass@1 performance across eight optimization modeling benchmarks. \textsc{SAGE} achieves the strongest results among open source models. It also surpasses its teacher model DeepSeek-R1, which is used during distillation, despite using far fewer parameters than the largest generalist LLMs. This result supports the effectiveness of strategy aware post training with solver feedback, which enables a compact model to acquire more reliable optimization modeling behavior.

Compared with online RL baselines such as SIRL-Qwen2.5-7B and StepORLM-Qwen3-8B, our model achieves comparable accuracy on Easy benchmarks and delivers substantially stronger performance on Complex benchmarks. Averaged over Complex benchmarks, \textsc{SAGE} improves Pass@1 by approximately \textbf{15.4\%}. These results indicate that making Modeling Strategy explicit is particularly beneficial for structurally complex problems.

The framework also generalizes across backbone scales. When applied to a smaller Qwen3-8B backbone, our method still achieves the best performance on the same challenging benchmarks. Detailed results are provided in Appendix~\ref{app:res_qwen3}.

\begin{figure}[t]
    \centering
    \includegraphics[width=\columnwidth]{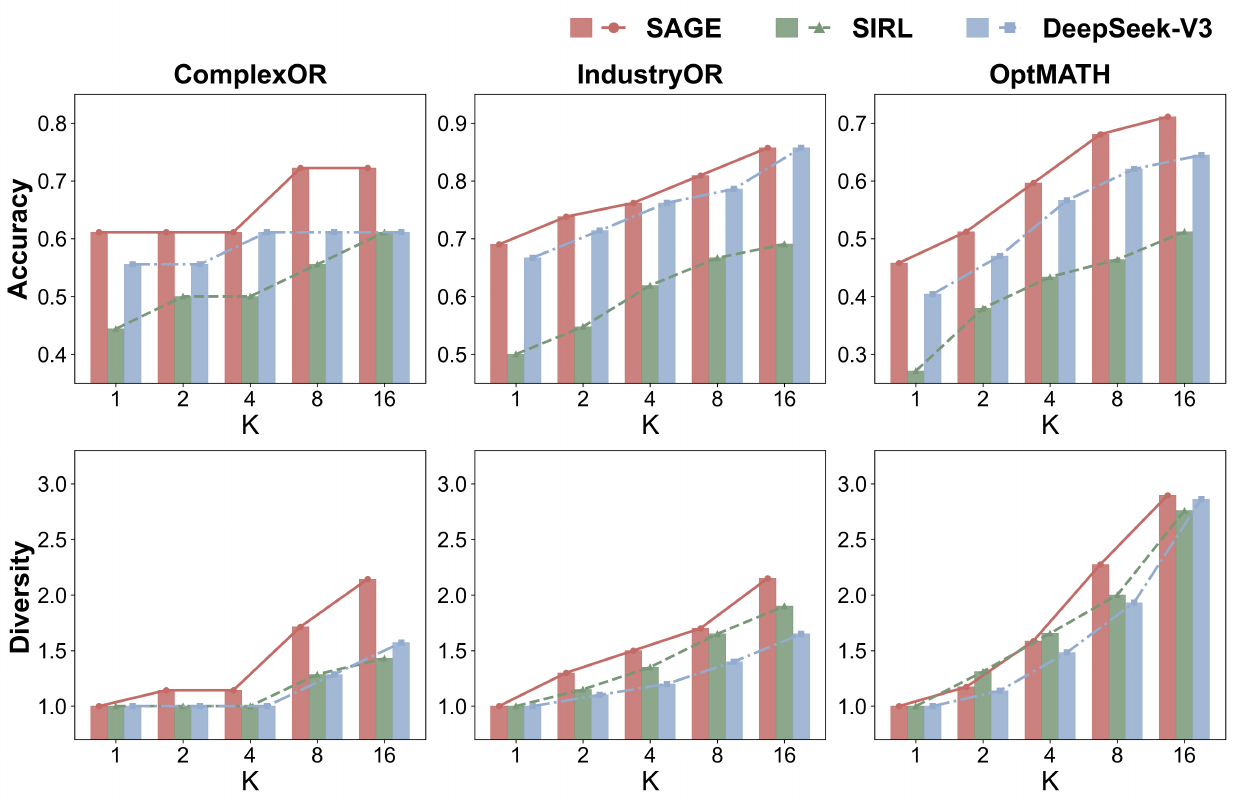}
    \caption{\textbf{Pass@K accuracy and modeling diversity.} Our method continues to discover more correct and more diverse formulations as $K$ increases.}
    \label{fig:passk}
\end{figure}

\subsection{Pass@K Performance}
We further evaluate Pass@K to measure modeling competence when multiple generations are allowed. This setting tests whether additional sampling enables the model to discover more correct solutions and more distinct valid formulations. We run this evaluation on three challenging benchmarks, \textit{CpxOR}, \textit{IndOR}, and \textit{OptM.}, and compare against two strong baselines, SIRL and DeepSeek-V3. These baselines achieve the highest average accuracy among open-source baselines in Table~\ref{tab:opt_results}, and all methods share the same Gurobi solver environment.

\paragraph{Accuracy.}
Across all three benchmarks, our model consistently achieves higher Pass@K accuracy than both baselines under every sampling budget, shown in figure~\ref{fig:passk}. At pass@16, our method reaches 0.72 on ComplexOR, 0.86 on IndustryOR, and 0.71 on OptMATH. As $K$ increases, the performance gap remains stable, which suggests that our approach produces correct formulations more reliably and benefits more effectively from additional sampling.

\begin{table}[t]
\centering
\small
\setlength{\tabcolsep}{8pt}
\renewcommand{\arraystretch}{1.1}
\caption{\textbf{Component level modeling diversity at pass@16.}
We report the average number of distinct decision variable, constraint, and objective per problem across multiple correct generations. Higher values indicate greater modeling diversity.
Higher values indicate greater modeling diversity.}
\label{tab:passk_diversity_components}
\begin{tabular}{l *{3}{S[table-format=1.2, detect-weight]}}
\toprule
Model & {Var. Types} & {Constr. Types} & {Obj. Types} \\
\midrule
\rowcolor{gray!10}
\textbf{SAGE (Ours)} & \textbf{2.33} & \textbf{2.31} & \textbf{2.08} \\
SIRL                & 1.80          & 1.91          & 1.74          \\
DeepSeek-V3         & 1.94          & 2.03          & 1.81          \\
\bottomrule
\end{tabular}
\end{table}

\paragraph{Diversity.}
To assess the ability to explore diverse valid formulations, we define a metric for modeling diversity. For each dataset $d$, we collect the subset of problems solved correctly by all compared methods at any Pass@K, denoted as $\mathcal{P}_d$. For a problem $p \in \mathcal{P}_d$, let $\mathcal{S}_p^K = \{s_{p,1}, \dots, s_{p,K}\}$ be the first $K$ generated solutions, and let $\phi(s_{p,i})$ denote the corresponding modeling representation. We define diversity as
\begin{equation}
\mathrm{Div}_d(K)
=
\frac{1}{|\mathcal{P}_d|}
\sum_{p \in \mathcal{P}_d}
\left|
\left\{
\phi(s_{p,i}) \;\middle|\;
s_{p,i} \text{ is correct},\; i \le K
\right\}
\right|.
\end{equation}
This metric reports, for each instance, the number of correct and non-equivalent formulations among the first $K$ generations, averaged over the benchmark. We treat two generated model as identical if they correspond to the same linear program (LP) formulation up to semantics-preserving rewrites (e.g., variable renaming and re-ordering). We use an LLM-based judge to compare the exported LP files and assess whether two formulations should be considered semantically equivalent for deduplication.

Our method also improves diversity of correct formulations, shown in figure~\ref{fig:passk}. Table~\ref{tab:passk_diversity_components} shows higher component level diversity, measured by the number of distinct variable designs, constraint structures, and objective formulations produced for the same problem across multiple correct generations. These results indicate that explicit Modeling Strategy improves correctness while expanding the range of valid formulation structures that the model can produce.

\begin{table*}[t]
\caption{\textbf{Ablation study of different training components.}
We report pass@1 accuracy (\%) on eight optimization benchmarks.
Blue and red arrows indicate performance improvements and degradations relative to the full training model, respectively.}
\centering
\label{tab:ablation}
\small
\setlength{\tabcolsep}{5pt}
\begin{tabular}{l c cccc cccc c}
\toprule
\multirow{2.5}{*}{Models} & \multirow{2.5}{*}{NL4OPT} & \multicolumn{2}{c}{MAMO} & \multirow{2.5}{*}{NLP4LP} & \multirow{2.5}{*}{CpxOR} & \multirow{2.5}{*}{IndOR} & \multirow{2.5}{*}{OptiB.} & \multirow{2.5}{*}{OptM.} & \multirow{2.5}{*}{\textbf{Avg.}} \\
\cmidrule(lr){3-4}
& & {Easy} & {Complex} & & & & & & \\
\midrule

Full Training
& 94.3
& 94.7
& 84.7
& 98.9
& 61.1
& 69.0
& 93.8
& 45.8 
& 80.3\\
\midrule

w/o RL
& 92.0 {\scriptsize\textcolor{red}{$\downarrow$2.3}}
& 89.5 {\scriptsize\textcolor{red}{$\downarrow$5.2}}
& 69.4 {\scriptsize\textcolor{red}{$\downarrow$15.3}}
& 94.9 {\scriptsize\textcolor{red}{$\downarrow$4.0}}
& 55.6 {\scriptsize\textcolor{red}{$\downarrow$5.5}}
& 57.1 {\scriptsize\textcolor{red}{$\downarrow$11.9}}
& 91.6 {\scriptsize\textcolor{red}{$\downarrow$2.2}}
& 42.2 {\scriptsize\textcolor{red}{$\downarrow$3.6}} 
& 74.0 {\scriptsize\textcolor{red}{$\downarrow$6.3}} \\

RL w/o template
& 93.0 {\scriptsize\textcolor{red}{$\downarrow$1.3}}
& 94.1 {\scriptsize\textcolor{red}{$\downarrow$0.6}}
& 77.5 {\scriptsize\textcolor{red}{$\downarrow$7.2}}
& 97.8 {\scriptsize\textcolor{red}{$\downarrow$1.1}}
& 61.1 {\scriptsize\textcolor{black}{$\downarrow$0.0}}
& 66.7 {\scriptsize\textcolor{red}{$\downarrow$2.3}}
& 93.1 {\scriptsize\textcolor{red}{$\downarrow$0.7}}
& 42.8 {\scriptsize\textcolor{red}{$\downarrow$3.0}} 
& 78.3 {\scriptsize\textcolor{red}{$\downarrow$2.0}} \\

RL w/o weighted
& 92.5 {\scriptsize\textcolor{red}{$\downarrow$1.8}}
& 94.7 {\scriptsize\textcolor{black}{$\downarrow$0.0}}
& 73.9 {\scriptsize\textcolor{red}{$\downarrow$10.8}}
& 96.6 {\scriptsize\textcolor{red}{$\downarrow$2.3}}
& 55.6 {\scriptsize\textcolor{red}{$\downarrow$5.5}}
& 64.3 {\scriptsize\textcolor{red}{$\downarrow$4.7}}
& 91.8 {\scriptsize\textcolor{red}{$\downarrow$2.0}}
& 44.0 {\scriptsize\textcolor{red}{$\downarrow$1.8}} 
& 76.7 {\scriptsize\textcolor{red}{$\downarrow$3.6}}\\

RL w/o eff-reward
& 92.0 {\scriptsize\textcolor{red}{$\downarrow$2.3}}
& 95.0 {\scriptsize\textcolor{blue}{$\uparrow$0.3}}
& 72.1 {\scriptsize\textcolor{red}{$\downarrow$12.6}}
& 96.1 {\scriptsize\textcolor{red}{$\downarrow$2.8}}
& 61.1 {\scriptsize\textcolor{black}{$\downarrow$0.0}}
& 66.7 {\scriptsize\textcolor{red}{$\downarrow$2.3}}
& 91.6 {\scriptsize\textcolor{red}{$\downarrow$2.2}}
& 44.6 {\scriptsize\textcolor{red}{$\downarrow$1.2}} 
& 77.4 {\scriptsize\textcolor{red}{$\downarrow$2.9}} \\
\bottomrule
\end{tabular}
\end{table*}

\subsection{Efficiency Performance}

\begin{figure}[t]
    \centering
    \includegraphics[width=\columnwidth]{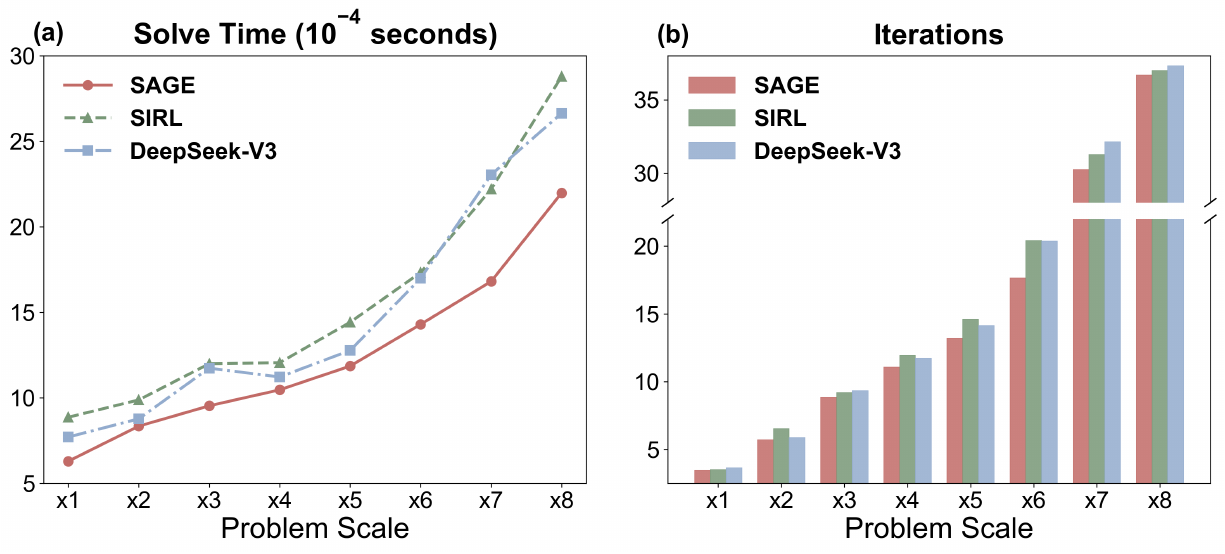}
    \caption{\textbf{Efficiency performance under increasing problem scale.} Our method yields lower solve time and fewer solver iterations, with larger gains on larger instances.}
    \label{fig:scale_efficiency}
\end{figure}

We examine whether the Modeling Strategy learned by our method improves solver efficiency as problem scale increases, beyond improving formulation correctness. We focus on ComplexOR because its benchmark design separates problem descriptions from numerical data, which enables controlled scaling of problem sizes.

To reduce the influence of correctness on efficiency comparisons, we select seven representative problems that can be solved stably by all compared methods. For each selected problem, we treat the original variable scale as the base and construct larger variants by linearly increasing the number of decision variables. For each scale, we generate five random data instances under the same structural constraints.

We construct the evaluation set by using DeepSeek-V3 to generate solver-executable code and executing it with Gurobi. We retain instances whose solver-returned solutions are self-consistent across multiple sampled generations as reference solutions \cite{wang2022self}. We evaluate solver efficiency using two indicators, the solve time reported by Gurobi to reflect computational cost and the number of solver iterations to reflect convergence complexity. For each scale, we report averages over instances that are solved correctly. More scaling details are provided in Appendix~\ref{app:scaling}.

Figure~\ref{fig:scale_efficiency} shows that, across all scales, our model produces formulations that solve faster and require fewer iterations than formulations produced by the baselines. The gap grows as scale increases, which suggests that the efficiency benefits induced by strategy aware modeling become more pronounced on larger instances.

\begin{table}[t]
\centering
\small
\setlength{\tabcolsep}{10pt}
\renewcommand{\arraystretch}{1.1}
\caption{\textbf{Structural complexity of generated formulations.}
We report the average number of decision variables (\#Vars), constraints (\#Constr.), and non zero coefficients (NNZ) at the largest problem scale ($\times 8$).
Lower values are better.}
\label{tab:efficiency_structure}
\begin{tabular}{l S[table-format=3.1, detect-weight] S[table-format=3.1, detect-weight] S[table-format=3.1, detect-weight]}
\toprule
Model & {\#Vars} & {\#Constr.} & {NNZ} \\
\midrule
\rowcolor{gray!10}
\textbf{SAGE (Ours)}  & 205.7          & \textbf{100.6} & \textbf{503.9} \\
SIRL                 & 197.8          & 117.2          & 543.3          \\
DeepSeek-V3          & \textbf{191.8} & 116.8          & 520.2          \\
\bottomrule
\end{tabular}
\end{table}

To understand the source of these efficiency gains, we analyze the structural complexity of generated formulations at the largest scale ($\times 8$), as summarized in Table~\ref{tab:efficiency_structure}. We report three indicators. \textit{\#Vars} is the number of decision variables. \textit{\#Constr.} is the number of constraints. \textit{NNZ} is the number of non-zero coefficients in the constraint matrix. Together they reflect the size and sparsity of the resulting LP.

As shown in Table~\ref{tab:efficiency_structure}, SAGE uses slightly more variables, but consistently yields fewer constraints and lower NNZ than both baselines. This matches a common OR trade-off (extended formulations): introducing auxiliary variables can simplify and sparsify the constraint system, reducing the linear-algebra workload that dominates solver time. Hence, the efficiency gains are better explained by a sparser constraint matrix rather than minimizing \textit{\#Vars}.

\subsection{Ablation Study}

We conduct ablation experiments to examine the contribution of the core components in our framework: (1) w/o RL: training solely with supervised fine-tuning without reinforcement learning, serving as the base policy to assess the impact of the RL stage; (2) RL w/o template: replaces the structured reasoning format with a plain instruction during the RL phase, where the model generates the formulation without explicitly outputting a modeling strategy; (3) RL w/o weighted: sets all token weights to one, effectively degenerating the objective into standard GRPO training; and (4) RL w/o efficiency: removes the efficiency term from the reward, optimizing solely for format and correctness.

Table~\ref{tab:ablation} summarizes the results. The w/o RL variant shows the largest overall degradation, demonstrating that reinforcement learning is essential beyond imitation. Both the structured reasoning template and the weighted loss play crucial roles, particularly on structurally complex benchmarks such as \textit{MAMOComplex}, \textit{IndOR}, and \textit{OptM.}. Removing the structured template results in an average accuracy drop of \textbf{4.2\%}, while removing the weighted loss leads to a larger decline of \textbf{5.8\%} across these datasets. This indicates that the structured template provides high-level organization for multi-stage reasoning, while the weighted loss enables more precise credit assignment to the modeling strategy segment, ensuring that key strategic decisions receive stronger optimization feedback during training.

\begin{table}[t]
\centering
\small
\setlength{\tabcolsep}{3pt} 
\renewcommand{\arraystretch}{1.1}
\caption{
\textbf{Ablation results on IndustryOR grouped by problem difficulty.}
We report execution rate (ER, \%) and accuracy (AC, \%) for Easy, Medium, and Hard subsets.
}
\label{tab:industry_ablation_difficulty}

\begin{tabular}{l @{\hspace{8pt}} *{6}{S[table-format=3.1, detect-weight]}}
\toprule
\multirow{2.5}{*}{Models} & \multicolumn{2}{c}{Easy} & \multicolumn{2}{c}{Medium} & \multicolumn{2}{c}{Hard} \\
\cmidrule(lr){2-3} \cmidrule(lr){4-5} \cmidrule(lr){6-7}
& {ER} & {AC} & {ER} & {AC} & {ER} & {AC} \\
\midrule

\rowcolor{gray!10}
\textbf{Full Training} & \textbf{100.0} & 77.3 & \textbf{100.0} & \textbf{75.0} & \textbf{91.7} & \textbf{50.0} \\
w/o RL               & 90.9  & 77.3 & 75.0  & 50.0 & 58.3 & 25.0 \\
RL w/o template      & 95.5  & \textbf{81.8} & 87.5  & \textbf{75.0} & 75.0 & 33.3 \\
RL w/o weighted      & \textbf{100.0} & 77.3 & 87.5  & 62.5 & 83.3 & 41.7 \\
RL w/o eff-reward    & 95.5  & \textbf{81.8} & \textbf{100.0} & \textbf{75.0} & 83.3 & 41.7 \\
\bottomrule
\end{tabular}
\vspace{-0.5cm}
\end{table}

Table~\ref{tab:industry_ablation_difficulty} further confirms this observation, showing that the performance gap between the full model and its ablated variants widens as the task difficulty increases. Especially, while \textit{RL w/o template} achieves comparable or even slightly higher performance on easier problems, its accuracy and execution rate degrade sharply on medium and hard subsets.

\section{Related Work}

\subsection{LLMs for Automated Optimization Formulation}
Large language models have been applied to automated optimization formulation and solving in operations research, leveraging their capabilities in mathematical reasoning and code generation~\cite{brown2020language, wei2022chain}. Early work such as NL4OPT shows that LLMs can translate natural language problem descriptions into linear optimization formulations by identifying decision variables, constraints, and objectives~\cite{ramamonjison2023nl4opt}. Subsequent pipeline based approaches integrate LLMs with optimization solvers to produce solver executable models, including AutoFormulation and OptiMUS~\cite{ahmaditeshnizi2024optimus, huang2025orlm}. In parallel, learning based methods fine tune LLMs on optimization specific datasets to improve formulation accuracy and executability, with representative examples including ORLM and LLMOpt~\cite{huang2025orlm, jiang2024llmopt}. More recent studies incorporate solver feedback during training or inference, including SIRL, which uses feasibility and optimality signals as rewards~\cite{chen2025sirl}, StepORLM, which introduces process level supervision for optimization reasoning~\cite{zhou2025steporlm}, and OR-R1, which applies test time policy optimization to refine formulations during inference~\cite{ding2025orr1}. Our work complements these directions by making Modeling Strategy explicit as a first class planning layer and optimizing it with solver verified data and strategy focused credit assignment.

\subsection{Post Training for LLM Reasoning}
Post training techniques are widely used to enhance reasoning capabilities. Prompting and few shot learning can improve performance without parameter updates, but they often remain limited on complex multi stage reasoning tasks~\cite{brown2020language}. Supervised fine tuning provides a systematic approach by training models to imitate high quality reasoning trajectories. Its effectiveness can be improved by rejection sampling, which enables large scale acquisition of high quality traces without additional human annotation~\cite{bai2022training}.

Beyond imitation, preference based methods such as Direct Preference Optimization provide an efficient framework for alignment, with performance that depends on the quality and diversity of preference data~\cite{rafailov2023dpo}. Reinforcement learning further enables direct optimization with verifiable feedback. Reinforcement learning from human feedback and policy gradient methods such as Proximal Policy Optimization have achieved strong results on reasoning benchmarks~\cite{christiano2017rlhf, schulman2017ppo, ouyang2022training, cobbe2021training}. Group Relative Policy Optimization is a critic free variant that improves efficiency by using relative reward comparisons within groups of sampled trajectories~\cite{shao2024deepseekmath}. Our Segment Weighted GRPO builds on GRPO and improves credit assignment in long horizon modeling by emphasizing the strategy segment, while also incorporating solver efficiency into the training objective.

\section{Conclusion}
We study strategy aware optimization modeling and propose \textsc{SAGE}, a framework that makes Modeling Strategy explicit in both data construction and post training. \textsc{SAGE} builds a solver verified multi strategy dataset and trains with supervised fine tuning followed by Segment Weighted GRPO using a composite reward over format compliance, correctness, and solver efficiency. This design enables the model to generate formulations that are correct and solver efficient. Experiments show improvements in pass@1 accuracy, pass@K behavior, modeling diversity, and solver efficiency across benchmarks. These results indicate that explicit Modeling Strategy is important for reliable and efficient automated optimization modeling.

\section*{Impact Statement}
This paper presents work whose goal is to advance the field of Machine
Learning. There are many potential societal consequences of our work, none
which we feel must be specifically highlighted here.

\bibliography{reference}
\bibliographystyle{icml2026}

\newpage
\appendix
\onecolumn

\section{Additional Experiment Details}

\subsection{Benchmark Details}
\label{app:benchmark}
Recent years have witnessed rapid progress in dataset construction for natural language optimization modeling, with a growing number of benchmarks and synthesis pipelines proposed to improve the ability of large language models to formulate and solve operations research problems. These datasets cover linear programming (LP), mixed-integer linear programming (MILP), and real-world industrial optimization scenarios. However, the increasing complexity also introduces non-negligible quality issues, including logical inconsistencies, underspecified problem statements, and incorrect reference solutions. A recent research survey systematically examined widely-used optimization datasets described in natural language, identifying and correcting a substantial portion of such issues through manual verification and revalidation \cite{xiao2025survey}. In our experiments, we strictly follow this cleaned and verified benchmark protocol, and all evaluations are conducted on the corrected subsets to ensure robustness and fairness. The number of problems within each dataset is detailed in Table~\ref{tab:benchmark_stats}.

Specifically, we evaluate our method on the following representative benchmarks:

\begin{itemize}[topsep=0pt, itemsep=3pt, parsep=0pt, partopsep=0pt]
    \item \textbf{NL4Opt.} 
    A benchmark consisting of natural language descriptions of linear programming problems, primarily focusing on the translation from textual problem statements to formal LP models. The problems span classical OR settings such as resource allocation and scheduling. 
    
    \item \textbf{MAMO.} 
    A modeling-oriented benchmark designed to assess formulation correctness rather than solution accuracy alone. It includes LP and MILP problems and is divided into EasyLP and ComplexLP subsets with distinct levels of structural difficulty. 

    \item \textbf{NLP4LP.} 
    A human-authored benchmark with fine-grained annotations, covering LP and MILP problems from classical OR domains such as knapsack, scheduling, and production planning. The dataset emphasizes executable and semantically correct formulations. 

    \item \textbf{ComplexOR.} 
    A small but challenging benchmark curated from advanced OR case studies, featuring conceptually complex mixed-integer formulations that often require hierarchical reasoning or time-indexed modeling.

    \item \textbf{IndustryOR.} 
    A real-world industrial benchmark comprising optimization problems from manufacturing, logistics, finance, and energy domains. The problems span multiple formulation types and are annotated by difficulty level.

    \item \textbf{OptiBench} 
    A large-scale dataset generated via reverse Socratic synthesis, where optimization demonstrations are transformed into natural language problem descriptions. Compared to earlier synthetic datasets, it provides higher-quality intermediate reasoning structures.

    \item \textbf{OptMATH.}
    A benchmark designed to overcome the limited diversity and difficulty of earlier optimization modeling datasets. It covers a wide range of optimization paradigms beyond linear formulations, including LP, MILP, nonlinear and second-order cone programming.
\end{itemize}

\begin{table}[h!]
\centering
\setlength{\tabcolsep}{3pt}
\caption{Summary of evaluation datasets and number of validated instances used in our experiments.}
\label{tab:benchmark_stats}
\begin{tabular}{lcccccccc}
\toprule
Dataset 
& NL4Opt 
& MAMOEasy
& MAMOComplex
& NLP4LP 
& ComplexOR 
& IndustryOR 
& OptiBench 
& OptMATH \\
\midrule
\# Instances 
& 213 
& 545 
& 111 
& 178 
& 18 
& 42 
& 403 
& 166 \\
\bottomrule
\end{tabular}
\end{table}

\subsection{Scaling Problem Details}
\label{app:scaling}

This section provides additional details on the construction of scaled optimization problems used in the efficiency performance evaluation.

\textbf{Problem Selection.}
To isolate solver efficiency from solution correctness, we restrict the evaluation to a subset of problems that can be solved stably and correctly by all compared methods.
Specifically, we select seven problems from the ComplexOR benchmark for which SIRL, DeepSeek-V3, and SAGE consistently produce correct and feasible solutions.

\textbf{Problem Types.}
The selected problems consist of seven representative instances drawn from the ComplexOR benchmark, covering a diverse set of classical operations research formulations. Specifically, the problems include: (i) a Car Selection Problem, (ii) a capacitated Transportation Problem, (iii) a Knapsack Problem, (iv) a Cutting Stock Problem, (v) a Diet Problem, (vi) a Project Assignment Problem, and (vii) an Aircraft Assignment Problem.
This diversity ensures that the efficiency evaluation is not biased toward a specific problem class or formulation type.

\textbf{Scaling Procedure.}
For each selected problem, the original instance is treated as the base scale. Larger instances are constructed by linearly expanding the key structural dimensions of the problem. Formally, for a given base size $b$ and scale factor $s$, the scaled size is defined as $\min(b \times s, b_{\max})$, where $b_{\max}$ is a problem-specific upper bound following the standard practices. Importantly, the underlying problem structure and constraint logic remain unchanged across scales.

\textbf{Instance Generation and Verification.}
For each problem and each scale factor $s \in \{1, \ldots, 8\}$, we generate five independent instances using a deterministic seed derived from the problem name and scale index.
All instances are constructed to be feasible by design. Each generated instance is further verified by an external solver before inclusion in the evaluation.
Overall solver efficiency metrics are reported by averaging over all correct solutions at each scale.

\begin{tcolorbox}[
    title=\textbf{Case: Scaling a Transportation Problem},
    colframe=blue!60!black,
    colback=blue!5,
    boxrule=1.0pt,
    arc=1mm,
    breakable
]

\textbf{Base Problem Statement}

Consider a transportation problem. Given a set of \texttt{Origins} and a set of \texttt{Destinations}, each origin $i$ has a $\texttt{Supply}_i$ and each destination $j$ has a $\texttt{Demand}_j$.
Shipping one unit of goods from origin $i$ to destination $j$ incurs a cost $\texttt{Rate}_{i,j}$ and is subject to an upper bound $\texttt{Limit}_{i,j}$.
The goal is to minimize the total transportation cost while satisfying all supply, demand, and capacity constraints.

\medskip
\textbf{Variable assignments (base scale)}

\texttt{supply = [20, 30]; demand = [30, 20]; rate = [[8, 6],[5, 10]]; limit  = [[15, 25],[25, 20]]}

\tcbline

\begin{algorithm}[H]
\caption{Transportation Problem Automated Scaling Procedure}
\label{alg:transport-scale}
\begin{algorithmic}[1]
\INPUT base instance sizes $|\mathcal{O}_0|$, $|\mathcal{D}_0|$ and scale factor $s$
\STATE Set $|\mathcal{O}| \leftarrow s \cdot |\mathcal{O}_0|$
\STATE Set $|\mathcal{D}| \leftarrow s \cdot |\mathcal{D}_0|$
\STATE Sample capacity limits for all origin--destination pairs
\STATE Sample transportation costs for all origin--destination pairs
\STATE Construct destination demands from column capacity totals
\STATE Construct origin supplies from row capacity totals
\STATE Adjust boundary rows or columns to enforce total balance
\OUTPUT a feasible scaled transportation instance
\end{algorithmic}
\end{algorithm}

\tcbline

\textbf{Scaled Problem at $\times 8$.}
At scale factor $s = 8$, the transportation problem expands from $2 \times 2$ to $16 \times 16$ origins and destinations, increasing the number of decision variables from $4$ to $256$.
The problem statement and constraint structure remain identical to the base formulation.

\medskip
\textbf{Variable assignments (at $\times 8$ scale)}
\begin{verbatim}
supply = [503, 604, 376, 434, 520, 454, ..., 458, 485] % 16 origins
demand = [420, 416, 538, 425, 434, 532, ..., 518, 496] % 16 destinations
rate   = [
           [ 6,  8,  9, 11, 13,  8, ...,  9,  7],
           [10, 15, 11, 14,  6,  4, ...,  7,  2],
           ...
           [ 5, 15, 12,  8,  5,  9, ...,  2, 10]
         ]                                          % 16 x 16 cost matrix
limit  = [
           [42, 53, 31, 48, 41, 57, ..., 57, 52],
           [31, 55, 26, 30, 35, 54, ..., 50, 59],
           ...
           [22, 29, 40, 40, 28, 44, ..., 56, 39]
         ]                                          % 16 x 16 capacity matrix
\end{verbatim}
\end{tcolorbox}

\subsection{Hyperparameter Settings}
\label{app:hyperparameter}

We report the main hyperparameter configurations used in both the supervised fine-tuning and reinforcement learning stages. 
All training experiments were conducted on a single compute node equipped with eight NVIDIA H100 (80 GB) GPUs. 
The selected hyperparameters were empirically tuned to ensure stable optimization, efficient convergence, and balanced reward scaling across stages. 
Table~\ref{tab:train_hyperparams} summarizes the complete training configurations.

\begin{table}[h!]
\centering
\renewcommand{\arraystretch}{1.05}
\setlength{\tabcolsep}{8pt}
\caption{Key hyperparameters for SFT and RL training.}
\label{tab:train_hyperparams}
\small
\begin{tabular}{llll}
\toprule
\textbf{Stage} & \textbf{Type} & \textbf{Parameter} & \textbf{Value} \\
\midrule

\multirow{4}{*}{\textbf{SFT}}
 & \multirow{4}{*}{\textbf{Data}}
 & Optimizer & AdamW \\
 &  & Learning rate & $1\times10^{-5}$ \\
 &  & Max sequence length & 8192 \\
 &  & Training epochs & 2 \\
\midrule

\multirow{14}{*}{\textbf{RL}}
 & \textbf{Algorithm} & Advantage estimator & GRPO \\
\cmidrule(lr){2-4}
 & \multirow{4}{*}{\textbf{Data}}
 & Batch size & 32 \\
 &  & Learning rate & $1\times10^{-6}$ \\
 &  & Max prompt length & 2048 \\
 &  & Max response length & 8192 \\
\cmidrule(lr){2-4}
 & \multirow{6}{*}{\textbf{Actor/Rollout}}
 & KL loss type & low\_var\_kl \\
 &  & KL loss coefficient & 0.001 \\
 &  & PPO mini batch size & 8 \\
 &  & PPO micro batch size per GPU & 4 \\
 &  & Rollout number & 8 \\
 &  & Training epochs & 2 \\
\cmidrule(lr){2-4}
 & \multirow{3}{*}{\textbf{Segment Weights}}
 & strategy & 2.0 \\
 &  & modeling & 1.5 \\
 &  & others & 1.0 \\
\bottomrule
\end{tabular}
\end{table}

\section{Additional Experiment Results}
\subsection{Main results of Qwen3-8B}
\label{app:res_qwen3}

Table~\ref{tab:qwen3_8b_results} reports the main results under the Qwen3-8B backbone, comparing our method with representative online RL baselines, including SIRL and StepORLM.
All methods are evaluated under the same Pass@1 accuracy metric.

Overall, SAGE-Qwen3-8B achieves the highest average performance among Qwen3-8B-based models.
While the performance on easier benchmarks is generally comparable across methods, the advantage of our approach becomes more pronounced on complex tasks.
In particular, SAGE-Qwen3-8B consistently outperforms prior methods on MAMO-Complex, ComplexOR, IndustryOR, and OptMATH, indicating stronger robustness in handling structurally complex and large-scale optimization problems.

These results are consistent with the main findings reported in the full-scale experiments, further demonstrating that the proposed strategy-aware reinforcement learning framework remains effective even under a fixed and relatively compact model backbone.

\begin{table*}[h!]
\centering
\caption{\textbf{Main results of Qwen3-8B models with Pass@1 accuracy (\%).}
The best results are highlighted in \textbf{bold} and the second-best results are \underline{underlined}.}
\label{tab:qwen3_8b_results}
\small
\setlength{\tabcolsep}{4pt}
\renewcommand{\arraystretch}{1.1}

\begin{tabular}{ll *{8}{S[table-format=2.1, detect-weight]} c}
\toprule
\multirow{2.5}{*}{Types} & \multirow{2.5}{*}{Models} 
& \multicolumn{4}{c}{Easy Tasks} 
& \multicolumn{4}{c}{Complex Tasks} 
& \multirow{2.5}{*}{Avg.} \\
\cmidrule(lr){3-6} \cmidrule(lr){7-10}
& 
& {NL4OPT} & {\makecell{MAMO\\Easy}} & {NLP4LP} & {OptiB.} 
& {\makecell{MAMO\\Cpx.}} & {CpxOR} & {IndOR} & {OptM.} & \\
\midrule

\multirow{3}{*}{Online-RL}
& SIRL-Q2.5-7B   
& \underline{94.8} & \textbf{98.0} & \underline{96.6} & \underline{89.6} 
& \underline{81.1} & 44.4 & 50.0 & \underline{27.1} & \underline{72.7} \\

& StepORLM-Q3-8B 
& \textbf{96.7} & \underline{95.4} & \textbf{97.2} & 88.6 
& 79.3 & \underline{50.0} & \underline{52.4} & 13.9 & 71.7 \\

\cmidrule(lr){2-11}
\rowcolor{hl}
\cellcolor{white} & \textbf{SAGE-Qwen3-8B (Ours)} 
& 93.9 & 95.0 & \underline{96.6} & \textbf{92.1} 
& \textbf{82.0} & \textbf{55.6} & \textbf{57.1} & \textbf{43.4} & \textbf{77.0} \\

\bottomrule
\end{tabular}
\end{table*}

\subsection{Case Study}
\paragraph{Case Study 1: Strategy Diversity with Increasing Pass@K.}
This case study illustrates how our model discovers increasingly diverse optimization modeling strategies as the sampling budget increases.
We consider a project assignment problem drawn from the ComplexOR benchmark and examine two fundamentally different optimization formulations generated by our model at different sampling stages.

The first formulation, produced at pass@1, adopts a compact continuous transportation-style modeling strategy, where working hours are treated as divisible flow variables directly allocated between people and projects. In contrast, a second formulation emerges only at pass@16, which follows a fundamentally different modeling paradigm by discretizing working hours into unit time slots and formulating the problem as a time-indexed 0–1 assignment model.
Although both formulations are mathematically valid and yield correct solutions, they differ substantially in their underlying modeling philosophy, variable granularity, constraint construction, and computational characteristics, as summarized in Table~\ref{tab:strategy_comparison}.

\begin{table*}[h!]
\centering
\small
\renewcommand{\arraystretch}{1.2}
\setlength{\tabcolsep}{6pt}

\caption{Comparison of Two Modeling Strategies Generated at Different Pass@K}
\label{tab:strategy_comparison}

\begin{tabular}{p{2.8cm} p{6.5cm} p{6.5cm}}
\toprule

& \textbf{Continuous Flow Model} (pass@1) 
& \textbf{Unit-Time 0--1 Assignment Model} (pass@16) \\
\midrule

\textbf{Modeling Strategy}
& Directly model hour flow between people and projects 
& Discretize hours into unit time slots and assign individually \\

\textbf{Model Type }
& Linear Programming (LP) 
& Integer Programming (IP / MILP) \\

\textbf{Decision Variables }
& $x_{ij}$: total hours from person $i$ to project $j$ 
& $x_{ijk}$: whether hour $k$ of person $i$ is assigned to project $j$ \\

\textbf{Variable Type }
& Continuous ($x_{ij} \ge 0$) 
& Binary ($x_{ijk} \in \{0,1\}$) \\

\textbf{Supply Constraints }
& $\sum_j x_{ij} = \text{Supply}_i$ 
& $\sum_j x_{ijk} = 1,\ \forall i,k$ \\

\textbf{Practical Efficiency }
& High, scalable 
& Low, variable explosion \\

\textbf{Compactness }
& Highly compact 
& Highly disaggregated \\
\bottomrule
\end{tabular}
\end{table*}

\begin{tcolorbox}[
    title=\textbf{Case 1: Diverse Optimization Modeling},
    colback=modelbg,
    colframe=modelframe,
    boxrule=1.0pt,
    arc=1mm,
    breakable
]

\textbf{Problem Statement}

Consider a project assignment problem. Given a set of people \texttt{People} and a set of projects \texttt{Projects}. Each person $i$ has a certain number of available hours $\mathrm{Supply}_i$ and each project $j$ requires a certain number of hours $\mathrm{Demand}_j$. The cost per hour of work for person $i$ on project $j$ is $\mathrm{Cost}_{i,j}$. Each person $i$ can contribute to project $j$ up to a maximum limit $\mathrm{Limit}_{i,j}$. The problem aims to minimize the total cost of assigning people to projects.
It is constrained that the total number of hours assigned from each person $i$ equals its supply and the total number of hours assigned to each project $j$ equals its demand.

\medskip
\textbf{Variable explanations}

\textbf{Args:}
\begin{itemize}[topsep=2pt, itemsep=3pt]
    \item \texttt{Supply}: available hours per person
    \item \texttt{Demand}: required hours per project
    \item \texttt{Cost}: cost per hour for each person--project pair
    \item \texttt{Limit}: maximum allowed contribution per pair
\end{itemize}

\textbf{Returns:}
\begin{itemize}[topsep=2pt, itemsep=3pt]
    \item \texttt{total\_cost}: minimized total assignment cost
\end{itemize}

\medskip
\textbf{Variable assignments}

\texttt{\{"supply": [8, 7], "demand": [5, 10], "cost": [[10, 20], [15, 25]], "limit": [[5, 6], [4, 7]]\}}

\tcbline

\textbf{Continuous Flow Model (pass@1)}

\begin{lstlisting}[style=pycode]
# Decision variables: x[i,j] >= 0 (hours)
x = model.addVars(len(supply), len(demand), lb=0.0)

# Objective
model.setObjective(
    gp.quicksum(cost[i][j] * x[i,j]
                for i in range(len(supply))
                for j in range(len(demand))),
    GRB.MINIMIZE
)

# Supply constraints
for i in range(len(supply)):
    model.addConstr(sum(x[i,j] for j in range(len(demand))) == supply[i])

# Demand constraints
for j in range(len(demand)):
    model.addConstr(sum(x[i,j] for i in range(len(supply))) == demand[j])

# Limit constraints
for i in range(len(supply)):
    for j in range(len(demand)):
        model.addConstr(x[i,j] <= limit[i][j])
\end{lstlisting}

\tcbline

\textbf{Unit-Time 0--1 Assignment Model (pass@16)}

\begin{lstlisting}[style=pycode]
# Binary variables x[i,j,k]
x = {}
for i in people:
    for j in projects:
        for k in range(supply[i]):
            x[i,j,k] = model.addVar(vtype=GRB.BINARY)

# Objective
model.setObjective(
    gp.quicksum(cost[i][j] * x[i,j,k]
                for i in people
                for j in projects
                for k in range(supply[i])),
    GRB.MINIMIZE
)

# Demand constraints
for j in projects:
    model.addConstr(sum(x[i,j,k] for i in people for k in range(supply[i]))
                    == demand[j])

# Limit constraints
for i in people:
    for j in projects:
        model.addConstr(sum(x[i,j,k] for k in range(supply[i])) <= limit[i][j])

# Assignment constraints
for i in people:
    for k in range(supply[i]):
        model.addConstr(sum(x[i,j,k] for j in projects) == 1)
\end{lstlisting}
\end{tcolorbox}

\paragraph{Case Study 2: Why Strategy-Aware Modeling Leads to Higher Solver Efficiency.}

In this case study, we investigate why the formulations generated by SAGE consistently achieve higher solver efficiency compared to strong baselines. We focus on a project assignment problem from the ComplexOR benchmark and compare two correct formulations: one generated by SAGE and the other by DeepSeek-V3.

Table~\ref{tab:efficiency_comparison} shows the solver efficiency details between these two formulation. Although both models encode the same optimization problem and yield valid solutions, the formulation produced by DeepSeek-V3 introduces redundant constraints and variables, resulting in a larger and less compact constraint matrix. In contrast, SAGE generates a more concise formulation that avoids unnecessary modeling artifacts. This structural difference directly translates into improved solver performance, as reflected in reduced model size and faster convergence.

\begin{table*}[h!]
\centering
\small
\setlength{\tabcolsep}{15pt}

\caption{Solver efficiency comparison between SAGE and DeepSeek-V3 formulations.}
\label{tab:efficiency_comparison}

\begin{tabular}{l c c c c c}
\toprule
\textbf{Model} 
& \textbf{\#Vars} 
& \textbf{\#Constr.} 
& \textbf{NNZ} 
& \textbf{Solver Time (s)} 
& \textbf{Total Iterations} \\
\midrule
\textbf{Compact Model (SAGE)} 
& 384 
& 40 
& 768 
& 0.0063 
& 29 \\

\textbf{Redundant Model (DS-V3)} 
& 384 
& 424 
& 1152 
& 0.0186 
& 29 \\
\bottomrule
\end{tabular}
\end{table*}

\begin{tcolorbox}[
    title=\textbf{Case 2: Higher Solver Efficiency Analysis},
    colback=modelbg,
    colframe=modelframe,
    boxrule=1.0pt,
    arc=1mm,
    breakable
]

\textbf{Problem Statement}  

The Aircraft Assignment Problem aims to assign aircraft to routes in order to minimize the total cost while satisfying demand constraints with available aircraft. The problem involves a set of aircraft and a set of routes. Given the costs of assigning an aircraft to a route, the objective is to minimize the total cost of the assignment. There are limited available aircraft, and it is constrained that the number of each aircraft allocated does not exceed its available number. 
Given the demand of each route and the capabilities (the largest number of people that can be carried) of an aircraft for a route, the demand constraint ensures that the total allocation for each route satisfies the demand. The problem seeks to find the most cost-effective assignment of aircraft to routes.

\medskip
\textbf{Variable explanations}

\textbf{Args:}
\begin{itemize}[topsep=2pt, itemsep=3pt]
    \item \texttt{availability}: list, availability of each aircraft
    \item \texttt{demand}: list, demand for each route
    \item \texttt{capabilities}: 2D list, capabilities of each aircraft for each route
    \item \texttt{costs}: 2D list, costs of assigning each aircraft to each route
\end{itemize}

\textbf{Returns:}
\begin{itemize}[topsep=2pt, itemsep=3pt]
    \item \texttt{min\_total\_cost}: float, the minimum total cost of the assignment
\end{itemize}

\tcbline

\textbf{Model 1: Compact Formulation (SAGE)}

\begin{lstlisting}[style=pycode]
# Define sets
num_aircraft = len(availability)
num_routes = len(demand)
I = range(num_aircraft)
J = range(num_routes)

# Decision variables
x = model.addVars(I, J, vtype=GRB.INTEGER, name="x")

# Objective
model.setObjective(
    gp.quicksum(costs[i][j] * x[i,j] for i in I for j in J),
    GRB.MINIMIZE
)

# Availability constraints
model.addConstrs(
    (gp.quicksum(x[i,j] for j in J) <= availability[i] for i in I),
    name="availability"
)

# Demand constraints
model.addConstrs(
    (gp.quicksum(capabilities[i][j] * x[i,j] for i in I) >= demand[j] for j in J),
    name="demand"
)
\end{lstlisting}

\tcbline

\textbf{Model 2: Redundant Formulation (DeepSeek-V3)}

\begin{lstlisting}[style=pycode]
# Sets
aircraft = range(len(availability))
routes = range(len(demand))

# Variables
x = model.addVars(aircraft, routes, vtype=GRB.INTEGER, name="x")

# Objective
model.setObjective(
    gp.quicksum(costs[i][j] * x[i,j] for i in aircraft for j in routes),
    GRB.MINIMIZE
)

# Availability constraints
for i in aircraft:
    model.addConstr(
        gp.quicksum(x[i,j] for j in routes) <= availability[i],
        f"availability_{i}"
    )

# Demand constraints
for j in routes:
    model.addConstr(
        gp.quicksum(capabilities[i][j] * x[i,j] for i in aircraft) >= demand[j],
        f"demand_{j}"
    )

# Non-negativity constraints
for i in aircraft:
    for j in routes:
        model.addConstr(x[i,j] >= 0, f"non_neg_{i}_{j}")
\end{lstlisting}
\end{tcolorbox}

\section{Additional Methodology Details}
\subsection{Efficiency Reward Design.}
\label{app:eff_reward}

To capture solver efficiency consistently across different optimization problem types, we design problem-specific efficiency metrics that reflect the computational cost observed during solver execution. 
For each correctly solved instance, we extract a scalar efficiency measure $M(y)$ from solver feedback and transform it into a bounded reward using:
\begin{equation}
R_{\text{efficiency}}(y) = 1 - \tanh\!\left(\frac{M(y)}{\alpha_{\text{eff}}}\right),
\end{equation}
where $\alpha_{\text{eff}}$ is a global scaling constant controlling the reward’s sensitivity to solver efficiency. 
A smaller $M(y)$ implies better solver performance, yielding a higher $R_{\text{efficiency}}(y)$. 
The definition of $M(y)$ depends on the underlying problem class, as detailed below.

\textbf{Linear Programs (LP).}
For linear programming problems, solver effort is primarily determined by the number of iterations required for convergence under simplex or barrier methods.
Hence, we define:
\begin{equation}
M_{\text{LP}}(y) = \text{IterationCount}(y).
\end{equation}
The efficiency reward then becomes:
\begin{equation}
R_{\text{efficiency}}^{\text{(LP)}}(y) = 1 - \tanh\!\left(\frac{\text{IterationCount}(y)}{\alpha_{\text{iter}}}\right).
\end{equation}
Here, $\alpha_{\text{iter}}$ is a scaling factor determining the solver iteration range considered “typical.” 
This design penalizes unnecessarily long convergence sequences while ensuring smooth gradient variation even for large-scale LPs.

\textbf{Mixed-Integer Linear Programs (MILP).}
For MILP problems, solver efficiency depends jointly on the tightness of continuous relaxations and the complexity of the branch-and-bound search process. 
Two solver-provided quantities are used:
the \textit{optimality gap}, measuring the relative distance between the incumbent and best-known bound, and 
the \textit{node count}, measuring the number of explored search tree nodes.
These are normalized and aggregated to form the unified metric:
\begin{equation}
M_{\text{MILP}}(y)
= 
\beta_{\text{gap}} \cdot \text{Gap}(y) + \beta_{\text{nodes}} \cdot \text{Nodes}(y),
\end{equation}
where $\beta_{\text{gap}}$ and $\beta_{\text{nodes}}$ control the relative contribution of the gap and node count. 
Both terms are first normalized by solver-scale statistics to balance their numerical ranges. 
The corresponding reward is given by:
\begin{equation}
R_{\text{efficiency}}^{\text{(MILP)}}(y) 
= 1 - \tanh\!\left(\frac{M_{\text{MILP}}(y)}{\alpha_{\text{eff}}}\right).
\end{equation}

This formulation encourages the model to generate formulations that improve solver relaxation tightness (lower gap) and reduce combinatorial branching (fewer nodes), both of which are well-established indicators of modeling efficiency in MILP research.

\textbf{Scaling and Weighting Factors.}
Table~\ref{tab:scaling_factors} summarizes all scaling and weighting factors used in the efficiency reward computation. 
The scaling factors $\alpha_{\text{iter}}, \alpha_{\text{gap}}, \alpha_{\text{nodes}}$ control the sensitivity of the reward 
to solver performance metrics, while the weighting factors $\beta_{\text{gap}}, \beta_{\text{nodes}}$ determine their 
relative contribution in mixed-integer optimization tasks. 
All parameters were empirically selected to ensure smooth reward shaping and numerical stability across scales.

\begin{table}[h]
\centering
\caption{Scaling and weighting factors used in the efficiency reward computation.}
\label{tab:scaling_factors}
\small
\renewcommand{\arraystretch}{1.1}
\setlength{\tabcolsep}{10pt}
\begin{tabular}{lccccc}
\toprule
& $\boldsymbol{\alpha_{\text{iter}}}$ & $\boldsymbol{\alpha_{\text{gap}}}$ & $\boldsymbol{\alpha_{\text{nodes}}}$ & $\boldsymbol{\beta_{\text{gap}}}$ & $\boldsymbol{\beta_{\text{nodes}}}$ \\
\midrule
\textbf{Value} & 15.0 & 1/3 & 5.0 & 0.5 & 0.5 \\
\bottomrule
\end{tabular}
\end{table}

\subsection{Prompt Templates}
\label{app:prompts}

\paragraph{Prompts for Data Generation.}
We use three prompt templates to generate diverse optimization modeling data in a structured pipeline. Each template serves a distinct role in producing strategy-diverse, solver-executable training samples. The first template is used to generate multiple modeling strategies for the same optimization problem. Given a selected strategy, the second template guides the model to produce a detailed step-by-step modeling reasoning and the corresponding Gurobi Python implementation. Finally, the third template is used for strategy de-duplication. Using an LLM-as-Judge, it compares the generated strategies and their solver code to identify and remove semantically redundant formulations that differ only superficially.

To assess the reliability of the LLM-as-Judge used for strategy de-duplication, we manually audited a random subset of 50 generated strategy pairs. On this subset, the judge’s decisions agreed with human judgments in approximately 87\% of cases. Most disagreements arose from borderline cases involving formulations with nearly identical mathematical structure, such as differences limited to variable renaming or constraint reordering. Importantly, all candidate formulations are solver-verified prior to de-duplication, thus the LLM-as-Judge is therefore only used to control redundancy among already correct and executable models, rather than to assess formulation correctness.

\begin{tcolorbox}[
    title=\textbf{\textcolor{white}{Modeling strategy generation prompt}},
    colframe=black,
    coltitle=black,
    boxrule=1.0pt,
    arc=1mm,
    breakable
]
You are a mathematical modeling expert.
Given the following optimization problem, propose \textbf{3 different modeling strategies} that could be used to formulate it.
Make sure the strategies differ not just in wording or variable names, but in modeling structure, such as using different variable types, constraint formulations, or problem decompositions.

\medskip
For each strategy, provide the following fields in the same structure:

------

\#\#\# Strategy X

Strategy Name: \textcolor{blue}{[Short name of the formulation]}

Description: \textcolor{blue}{[The summary of the approach]}

------

\medskip
Do not write equations or code. Keep the output structured and machine-readable.

Here is the problem:

\textcolor{red}{\{\{Question\}\}}
\end{tcolorbox}

\begin{tcolorbox}[
    title=\textbf{\textcolor{white}{Modeling reasoning \& Gurobi code generation prompt}},
    colframe=black,
    colback=black!3,
    boxrule=1.0pt,
    arc=1mm,
    breakable
]
You are a mathematical modeling expert. Given the following modeling strategy for an optimization problem, please do the following:

\begin{enumerate}[topsep=0pt, itemsep=2pt, parsep=0pt, partopsep=5pt]
    \item Write a detailed \textbf{step-by-step modeling reasoning} explaining how to build the model according to this strategy:
    \begin{itemize}[topsep=0pt, itemsep=2pt, parsep=0pt, partopsep=5pt]
        \item What sets are defined?
        \item What parameters are introduced?
        \item How are the decision variables designed?
        \item What is the objective function?
        \item What are the key constraints?
    \end{itemize}
    \item Then, write the full \textbf{Gurobi Python} code that implements this formulation.
\end{enumerate}

\#\#\# Strategy Description:

\textcolor{red}{\{\{Strategy\}\}}

\medskip
Only output the Python code inside a code block:

\begin{verbatim}
```python
# ... your final code ...
\end{verbatim}

\medskip
Here is the problem:

\textcolor{red}{\{\{Question\}\}}
\end{tcolorbox}

\begin{tcolorbox}[
    title=\textbf{\textcolor{white}{Strategy de-duplication prompt}},
    colframe=black,
    colback=black!3,
    boxrule=1.0pt,
    arc=1mm,
    breakable
]

\textbf{SYSTEM:}
You are an expert in combinatorial optimization and mathematical modeling.
Your task is to determine whether the given strategies represent fundamentally different modeling paradigms.
Judge differences based on the following aspects:
\begin{itemize}[topsep=0pt, itemsep=2pt, parsep=0pt, partopsep=5pt]
    \item Variable definitions (binary, continuous, state-based, etc.)
    \item Objective formulation (value maximization, cost minimization, etc.)
    \item Constraints (linear, logical, recursive, etc.)
    \item Solution method (e.g., MIP, Dynamic Programming, Branch and Bound, Heuristic, Reinforcement Learning, etc.)
\end{itemize}
Two strategies are different if they differ in any of these aspects, even if they share similar wording or target the same problem.Return your answer in \textbf{STRICT JSON format} only.

\medskip
Your response must be a single valid JSON object with the exact structure:
\textcolor{blue}{\{"verdict": "different" or "similar", "similar\_ids": [list of ids if similar, otherwise empty list]\}}

\medskip
Do not include any text, explanation, markdown, or additional formatting outside the JSON braces.

\medskip
\textbf{USER:}
Here is the Strategy and its Gurobi code:
\textcolor{red}{\{\{Strategies\}\}} + \textcolor{red}{\{\{Gurobi Code\}\}}

\end{tcolorbox}

\paragraph{System Prompts and LLM Response}
These prompts are used during both the RL training stage and the inference stage to guide the model to generate a complete optimization formulation and executable solver code in a fixed and consistent format.
They enforce a standardized reasoning and output structure, which is essential for stable policy optimization, solver-based reward computation, and reliable evaluation.

In addition, we provide an example illustrating the input and output of this template on a representative optimization problem from the IndustryOR dataset.
The example shows how a natural language problem description is transformed into a structured reasoning trace and corresponding solver code under the strategy-aware format enforced by the system prompt.







\begin{tcolorbox}[
    title=\textbf{\textcolor{white}{Strategy-aware system prompt}},
    colframe=black,
    colback=black!3,
    boxrule=1.0pt,
    arc=1mm,
    breakable
]

\textbf{SYSTEM:}
You are an expert in mathematical optimization and operations research.
Your goal is to design a globally consistent mathematical model and implement it in Gurobi Python to solve the given problem.
Your reasoning must inside 
\textcolor{violet}{\texttt{<think>}}and 
\textcolor{violet}{\texttt{</think>}}
 tags and follow the strict format:

\medskip
\textcolor{violet}{\texttt{<think>}}

\textcolor{orange}{\texttt{<strategy>}}
\begin{enumerate}[topsep=0pt, itemsep=2pt, parsep=0pt, partopsep=5pt]
    \item Paradigm Selection: Identify the problem class (e.g., TSP, Scheduling, Facility Location) and choose \textbf{the most robust and efficient} modeling paradigm (e.g., Time-indexed, Flow-based, Big-M).
    \item Decision Variables: Define the core variables (names, indices, categories) that serve as the foundation.
    \item Constraint Logic: Identify key couplings and explain how the defined variables interact to enforce complex constraints.
\end{enumerate}
\textcolor{orange}{\texttt{</strategy>}}

\medskip
\textcolor{blue}{\texttt{<modeling>}}

Formulate the model following the design:
Sets, Parameters, Variables, Objective, Constraints.
Ensure every segment must match the strategy.

\textcolor{blue}{\texttt{</modeling>}}

\medskip
\textcolor{green!60!black}{\texttt{<check>}}
\begin{enumerate}[topsep=0pt, itemsep=2pt, parsep=0pt, partopsep=5pt]
    \item Consistency Check: Does the model strictly follow the selected strategy?
    \item Logic Check: Are the dependencies between variables and constraints logically sound?
    \item Dimension Check: Do the variable indices in the code match the constraint loops?
    \item Efficiency Check: Is the formulation compact and free of redundant variables?
\end{enumerate}
\textcolor{green!60!black}{\texttt{</check>}}

\textcolor{violet}{\texttt{</think>}}

\medskip
Output the final Python code inside a code block:
\begin{verbatim}
```python
# ... your final code ...
\end{verbatim}

\textbf{USER:}
Here is the problem:

\textcolor{red}{\{\{Question\}\}}

\end{tcolorbox}

\begin{tcolorbox}[
    title=\textbf{\textcolor{white}{Example Problem}},
    colframe=black,
    colback=black!3,
    boxrule=1.0pt,
    arc=1mm,
    breakable
]
\textbf{\# Question:}

The Zhang family has 6 children, Harry, Hermione, Ron, Fred, George, and Ginny. The cost of taking Harry is \$1200, Hermione is \$1650, Ron is \$750, Fred is \$800, George is \$800, and Ginny is \$1500. Which children should the couple take to minimize the total cost of taking the children?

\medskip
They can take a maximum of 4 children on the upcoming trip.

\medskip
Ginny is the youngest, so the Zhang family will definitely take her.

\medskip
If the couple takes Harry, they will not take Fred because Harry doesn't get along with him.

\medskip
If the couple takes Harry, they will not take George because Harry doesn't get along with him.

\medskip
If they take George, they must also take Fred.

\medskip
If they take George, they must also take Hermione.

\medskip
Although this will cost them a lot of money, the Zhang family has decided to take at least three children.

\end{tcolorbox}

\begin{tcolorbox}[
    title=\textbf{\textcolor{white}{Example Response: Reasoning \& Solver Code}},
    colframe=black,
    colback=black!3,
    boxrule=1.0pt,
    arc=1mm,
    breakable
]

\texttt{<think>}

\texttt{<strategy>}

\medskip
\begin{enumerate}[topsep=0pt, itemsep=2pt, parsep=0pt, partopsep=0pt]
    \item \textbf{Paradigm Selection}: This is a knapsack problem with additional constraints. We use binary decision variables for each child (1 if taken, 0 otherwise).
    \item \textbf{Decision Variables}:
    \begin{itemize}
        \item Let \texttt{x[i]} be a binary variable for each child $i \in \{\texttt{Harry}, \texttt{Hermione}, \texttt{Ron}, \texttt{Fred}, \texttt{George}, \texttt{Ginny}\}$.
    \end{itemize}
    \item \textbf{Constraint Logic}:
    \begin{itemize}
        \item Ginny must be taken: \texttt{x['Ginny'] = 1}.
        \item Maximum of 4 children: \texttt{sum(x[i] for i in children) <= 4}.
        \item Minimum of 3 children: \texttt{sum(x[i] for i in children) >= 3}.
        \item If Harry is taken, then Fred and George cannot be taken:
        \begin{itemize}
            \item \texttt{x['Harry'] + x['Fred'] <= 1}
            \item \texttt{x['Harry'] + x['George'] <= 1}
        \end{itemize}
        \item If George is taken, then Fred and Hermione must be taken:
        \begin{itemize}
            \item \texttt{x['George'] <= x['Fred']}
            \item \texttt{x['George'] <= x['Hermione']}
        \end{itemize}
        \item Note: The cost of each child is given as parameters.
    \end{itemize}
\end{enumerate}

\medskip
\texttt{</strategy>}

\medskip
\texttt{<modeling>}

\begin{itemize}[topsep=0pt, itemsep=2pt, parsep=0pt, partopsep=0pt]
    \item \textbf{Sets}:
    \begin{itemize}
        \item Children: \{\texttt{Harry}, \texttt{Hermione}, \texttt{Ron}, \texttt{Fred}, \texttt{George}, \texttt{Ginny}\}
    \end{itemize}

    \item \textbf{Parameters}:
    \begin{itemize}
        \item \texttt{cost}: dictionary mapping child to cost
        \item \texttt{max\_children} = 4
        \item \texttt{min\_children} = 3
    \end{itemize}

    \item \textbf{Variables}:
    \begin{itemize}
        \item \texttt{x[i]} for each child $i$: binary (1 if taken, 0 otherwise)
    \end{itemize}

    \item \textbf{Objective}:
    \begin{itemize}
        \item Minimize total cost: \texttt{sum(cost[i] * x[i] for i in children)}
    \end{itemize}

    \item \textbf{Constraints}:
    \begin{itemize}
        \item \texttt{x['Ginny'] == 1}
        \item \texttt{sum(x[i] for i in children) <= max\_children}
        \item \texttt{sum(x[i] for i in children) >= min\_children}
        \item \texttt{x['Harry'] + x['Fred'] <= 1}
        \item \texttt{x['Harry'] + x['George'] <= 1}
        \item \texttt{x['George'] <= x['Fred']}
        \item \texttt{x['George'] <= x['Hermione']}
    \end{itemize}
\end{itemize}

\medskip
\texttt{</modeling>}

\medskip
\texttt{<check>}

\begin{enumerate}[topsep=0pt, itemsep=2pt, parsep=0pt, partopsep=0pt]
    \item Consistency: The model uses binary variables and all constraints are linear.
    \item Logic: All constraints are logically derived from the problem description.
    \item Dimensions: Variable indices match the children set.
    \item Efficiency: The formulation is compact with no redundant variables.
\end{enumerate}

\medskip
\texttt{</check>}

\medskip
Now, the code:

\medskip
\texttt{</think>}

\begin{lstlisting}[style=pycode]
import gurobipy as gp
from gurobipy import GRB

# Create model
model = gp.Model("ZhangFamilyTrip")

# Children and costs
children = ['Harry', 'Hermione', 'Ron', 'Fred', 'George', 'Ginny']
costs = {
    'Harry': 1200,
    'Hermione': 1650,
    'Ron': 750,
    'Fred': 800,
    'George': 800,
    'Ginny': 1500
}

# Decision variables: binary for each child
x = model.addVars(children, vtype=GRB.BINARY, name="take")

# Objective: minimize total cost
model.setObjective(
    gp.quicksum(costs[i] * x[i] for i in children),
    GRB.MINIMIZE
)

# Constraints
model.addConstr(x['Ginny'] == 1, "Ginny_must_be_taken")
model.addConstr(gp.quicksum(x[i] for i in children) <= 4, "max_four")
model.addConstr(gp.quicksum(x[i] for i in children) >= 3, "min_three")

# Harry conflicts
model.addConstr(x['Harry'] + x['Fred'] <= 1, "Harry_Fred_conflict")
model.addConstr(x['Harry'] + x['George'] <= 1, "Harry_George_conflict")

# George dependencies
model.addConstr(x['George'] <= x['Fred'], "George_requires_Fred")
model.addConstr(x['George'] <= x['Hermione'], "George_requires_Hermione")

# Optimize
model.optimize()
\end{lstlisting}
\end{tcolorbox}

\end{document}